%% file: iclr2026_conference.tex
\documentclass{article} 
\usepackage{iclr2026_conference,times}

\input{math_commands.tex}

\usepackage[hidelinks, colorlinks]{hyperref}
\usepackage{url}
\usepackage{graphicx} 
\usepackage{booktabs} 
\usepackage{float}

\title{Implicit Statistical Inference in Transformers: Approximating Likelihood-Ratio Tests In-Context}

\author{Faris Chaudhry \& Siddhant Gadkari\\
Department of Computer Science\\
Imperial College London\\
\texttt{\{fc522,svg21\}@imperial.ac.uk} \\
}

\iclrfinalcopy
\begin{document}

\maketitle

\begin{abstract}
In-context learning (ICL) allows Transformers to adapt to novel tasks without weight updates, yet the underlying algorithms remain poorly understood. We adopt a statistical decision-theoretic perspective by investigating simple binary hypothesis testing, where the optimal policy is determined by the likelihood-ratio test. Notably, this setup provides a mathematically rigorous setting for mechanistic interpretability where the target algorithmic ground truth is known. By training Transformers on tasks requiring distinct geometries (linear shifted means vs. nonlinear variance estimation), we demonstrate that the models approximate the Bayes-optimal sufficient statistics from context up to some monotonic transformation, matching the performance of an ideal oracle estimator in nonlinear regimes. Leveraging this analytical ground truth, mechanistic analysis via logit lens and circuit alignment suggests that the model does not rely on a fixed kernel smoothing heuristic. Instead, it appears to adapt the point at which decisions become linearly decodable: exhibiting patterns consistent with a voting-style ensemble for linear tasks while utilizing a deeper sequential computation for nonlinear tasks. These findings suggest that ICL emerges from the construction of task-adaptive statistical estimators rather than simple similarity matching.
\end{abstract}

\section{Introduction}

In-context learning (ICL) refers to the remarkable ability of models (particularly Transformers) to adapt to novel tasks at inference time using only a finite context of input-output examples, without explicit parameter updates~\citep{brown2020language, vaswani2017attention}. While ICL is now a standard capability of large language models, its underlying algorithmic mechanism remains a subject of debate. Does the model merely retrieve and average similar examples, or does it construct a principled learning algorithm on the fly?

Recent work in controlled synthetic environments has demonstrated that Transformers can recover classical algorithms (such as linear regression, decision trees, and automata) purely from context~\citep{garg2022can, zhang2023trained}. These findings suggest that ICL may implement statistically optimal procedures when the task structure allows. However, existing analyses often focus on regression problems with fixed functional forms, emphasizing asymptotic convergence rather than the precise nature of the decision rule applied at the level of individual episodes.

In this work, we adopt a statistical decision-theoretic perspective. We study ICL in binary hypothesis testing, a fundamental framework where optimal decision rules are fully characterized by the Neyman-Pearson lemma~\citep{lehmann2005testing}. For simple hypotheses, the log-likelihood ratio (LLR) is a minimal sufficient statistic, and any Bayes-optimal decision rule must be a monotone function of it. This provides a sharp notion of optimality and identifiability: recovering the LLR up to a monotone (or affine) transformation is both necessary and sufficient for optimal prediction. More importantly, \emph{this establishes a testbed for mechanistic interpretability where the ground truth is known}, addressing a known challenge in mechanistic interpretability~\citep{sharkey2025openproblems}.

By training Transformers on dynamic discrimination tasks where the optimal statistic varies across episodes (e.g., linear vs.\ quadratic), we test whether the model learns to infer and apply the appropriate sufficient statistic from context alone, rather than relying on fixed similarity heuristics. We view this work as a first step toward a broader decision-theoretic understanding of ICL.

\subsection{Related Work}

\paragraph{ICL as implicit inference.}
A growing body of literature interprets ICL as a form of implicit Bayesian inference.~\citet{xie2022implicit} propose that ICL can be modeled as Bayesian inference over a hidden variable concept space, while~\citet{li2023transformers} and~\citet{zhang2023trained} demonstrate that Transformers can approximate posterior predictive distributions for specific function classes. Closest to our work,~\citet{bai2023transformers} analyze Transformers as statisticians in the context of Markov chains, finding that they can approach Bayes-optimal error rates. We extend this perspective by explicitly characterizing the geometry of the decision boundary (linear vs.\ quadratic) and linking the model's internal representations to the Neyman-Pearson optimal statistic.

\paragraph{Algorithmic induction and optimization.}
An alternative perspective views ICL as an optimization process.,~\citet{akyurek2022what}, ~\citet{dai2022why}, and~\citet{vonOswald2023transformers} have argued that self-attention layers can implement steps of gradient descent (GD) during the forward pass. While the ``ICL as GD'' hypothesis explains how models improve with more examples, it does not explicitly guarantee statistical optimality in discriminative settings. Our work complements this by focusing on the objective of the induced algorithm: regardless of whether the mechanism resembles GD or exact inference, we ask if it produces the sufficient statistic required for the likelihood-ratio test.

\paragraph{Mechanistic interpretability and task vectors.}
Finally, our analysis draws on mechanistic interpretability to explain how these statistics are computed~\citep{elhage2021mathematical, nanda2023progress}.~\citet{olsson2022context} identified induction heads as a primary circuit for copying patterns in ICL. More recently,~\citet{hendel2023incontext} and~\citet{todd2024function} have proposed that Transformers compress the context into function vectors or task vectors that modulate downstream processing. This aligns with our finding that the attention mechanism acts as a ``neural statistician'' of sorts~\citep{edwards2016towards}, compressing the context dataset into a single sufficient statistic (e.g., a mean vector or energy scalar) that determines the downstream decision rule.

\section{Problem Setup: Dynamic Statistical Discrimination}

We study ICL in the setting of binary hypothesis testing with task parameters that vary across episodes (i.e., independent task instances consisting of a context set and a query drawn from a shared latent task). Let $\Phi$ denote a family of binary classification tasks, where each task $\phi \in \Phi$ specifies two class-conditional distributions $(p_\phi(x \mid H_0), p_\phi(x \mid H_1))$ and an associated label space $y \in \{0,1\}$. In each episode, we sample task parameters $\phi \sim p(\Phi)$ and generate a context dataset $C = \{(x_i, y_i)\}_{i=1}^N$ where $y_i \sim \mathrm{Bernoulli}(1/2)$ and $x_i \sim p_\phi(x \mid H_{y_i})$. A query point $(x_q, y_q)$ is drawn from the same task distribution. A Transformer model $f_\theta$ is trained to predict the label (source distribution) $y_q$ given $(x_q, C)$ by minimizing the binary cross-entropy (BCE) loss:
\begin{equation}
    \mathcal{L} = -\mathbb{E}_{\phi \sim p(\Phi)} \mathbb{E}_{C, x_q} \left[ y_q \log f_\theta(x_q, C) + (1-y_q) \log (1 - f_\theta(x_q, C)) \right].
\end{equation}

Minimizing BCE is equivalent to estimating the posterior probability $p(y_q = 1 \mid x_q, C)$. The logit of the Bayes-optimal predictor satisfies
\begin{equation}
\log \frac{p(y_q = 1 \mid x_q, C)}{p(y_q = 0 \mid x_q, C)}
= \mathrm{LLR}(x_q; \phi) + \log \frac{\pi_1}{\pi_0},
\end{equation}
where $\pi_1, \pi_0$ denote the class priors. Thus, under BCE training, the Bayes-optimal internal decision statistic is identifiable up to an affine transformation of the LLR. 

Conditioned on the context dataset $C$, each episode induces a simple binary hypothesis testing problem between $H_0$ and $H_1$. By the Neyman-Pearson lemma, the likelihood-ratio test
\[
\frac{p(x_q \mid H_1, C)}{p(x_q \mid H_0, C)}
\]
is the uniformly most powerful decision rule, and any Bayes-optimal classifier must implement a statistic that is monotone in the corresponding log-likelihood ratio. Consequently, recovery of the LLR up to an affine transformation is both necessary and sufficient for optimal in-context prediction under BCE training.

To test whether Transformers rely on simple heuristics or perform optimal, context-dependent statistical inference, we design two Gaussian discrimination tasks with differing optimal statistics.

\paragraph{Task A: Shifted Mean Discrimination (Linear Regime).}
We sample a discriminative direction $\mu \sim \mathrm{Unif}(\mathbb{S}^{d-1})$ and a shift $k \sim \mathcal{N}(0, \sigma_k^2 I)$. The class-conditional distributions are
\begin{equation}
    H_0: x \sim \mathcal{N}(-\mu + k, I), \quad
    H_1: x \sim \mathcal{N}(\mu + k, I).
\end{equation}
The optimal decision boundary is linear but not centered at the origin. The sufficient statistic is the shifted projection $S(x) = \mu^\top (x - k)$, requiring the model to infer both $\mu$ and $k$ from the context. This task probes whether the model can dynamically estimate local centroids and perform linear discrimination. Static models that assume fixed centering fail on this task.

\paragraph{Task B: Variance Discrimination (Nonlinear Regime).}
We sample two variances $\sigma_0, \sigma_1 \sim \mathrm{Unif}[0.5, 3.0]$ and fix the mean at zero. The distributions are
\begin{equation}
    H_0: x \sim \mathcal{N}(0, \sigma_0^2 I), \quad
    H_1: x \sim \mathcal{N}(0, \sigma_1^2 I).
\end{equation}
Since the class means coincide, dot-product similarity is uninformative. The optimal decision statistic depends on the quadratic energy $\|x\|^2$, with the sign determined by the relative ordering of $(\sigma_0, \sigma_1)$. This task tests whether the model can adapt its internal geometry from linear projections to norm-based estimation.

\section{Approximation of the LLR}

\subsection{Recovery of Optimality}

To quantify the model's ability to recover the sufficient statistic, we compare its in-context accuracy against a theoretical Bayes-optimal classifier. The oracle computes the exact log-likelihood ratio using the ground-truth task parameters $(\mu, k, \sigma)$, representing the theoretical performance ceiling.

\begin{figure}[t]
    \centering
    \includegraphics[width=0.9\linewidth]{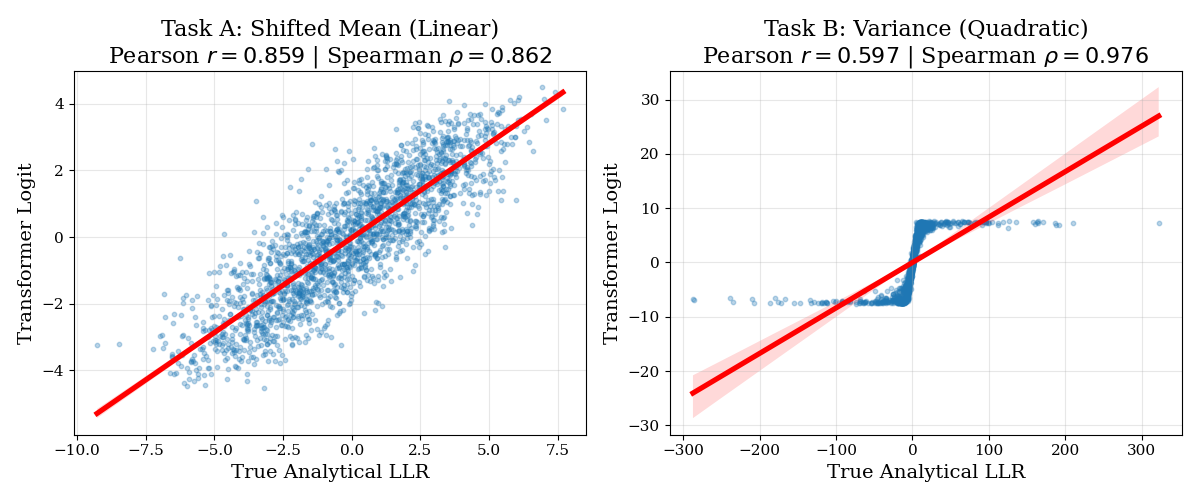}
    \caption{\textbf{Approximation of the LLR.} Regression of the Transformer's output logits against the true analytical LLR for validation episodes. \textbf{(Left) Task A:} The model exhibits a strong linear correlation ($r=0.859$), indicating it approximates the affine sufficient statistic $\mu^\top(x-k)$. \textbf{(Right) Task B:} The model achieves near-perfect rank correlation ($\rho=0.976$), effectively recovering the quadratic sufficient statistic $\|x\|^2$ up to a monotone transform. The sigmoidal shape suggests the model has learned a calibrated probability mapping, saturating for high-confidence inputs while preserving the optimal decision ordering.}
    \label{fig:logit-llr-regression}
\end{figure}

In the nonlinear variance task (Task B), the model achieves an accuracy of $83.0 \pm 0.5\%$, effectively matching the oracle performance of $84.0 \pm 1.0\%$. While the model's raw logits do not linearly track the analytical LLR (Pearson $r=0.60$), they achieve near-perfect rank alignment (Spearman $\rho=0.98$). This indicates that the model has successfully recovered the ordering induced by the quadratic sufficient statistic $\|x\|^2$, but maps it through a nonlinear calibration function (Figure \ref{fig:logit-llr-regression}).

In the linear shifted mean task (Task A), the model achieves $78.3 \pm 0.3\%$. While discriminative, it remains below the oracle accuracy of $84.6 \pm 1.0\%$, leaving an optimality gap of approximately $6.3\%$. This discrepancy is reflected in the regression analysis, which shows a noisy linear approximation ($r=0.86$) rather than the clean functional relationship observed in Task B. This suggests that instead of performing exact symbolic inference, the model implements some approximation. We verify this hypothesis in Appendix~\ref{app:ood_regression} by evaluating the model on OOD tasks with significantly larger nuisance shifts ($\sigma_k=9.0$). Under these conditions, the correlation with the true LLR degrades to $r=0.567$, demonstrating that the learned decision rule is a local approximation calibrated to the training support rather than an exact symbolic recovery. Nonetheless, the model does eventually begin to generalize OOD, exhibiting a delayed rise in validation accuracy characteristic of partial grokking.

\subsection{Ablations and Failure Modes}

We isolate the necessary components for in-context learning by modifying the architecture and data structure, as detailed in Table~\ref{tab:condensed_ablations}. Comprehensive results for all experimental conditions are provided in Appendix~\ref{app:full_ablation_results}.

\begin{table}[h]
\centering
\small
\caption{\textbf{Key Ablations (Task A).} We test the necessity of specific architectural features. \textbf{1) Permutation Invariance:} Removing positional encodings (\texttt{NoPos}) has negligible impact, confirming the model treats the context as a set rather than a sequence. \textbf{2) Learned Metric:} Freezing attention weights (\texttt{FrozenQK}) destroys performance, indicating the model must learn a task-specific similarity metric. \textbf{3) Supervision:} Shuffling labels (\texttt{ShuffledLabels}) causes collapse to random chance, ruling out unsupervised clustering heuristics.}
\label{tab:condensed_ablations}
\begin{tabular}{lcc}
\toprule
\textbf{Model Variant} & \textbf{Validation Accuracy} & \textbf{Implication} \\
\midrule
\texttt{Regular (Baseline)} & $78.3 \pm 0.3\%$ & --- \\
\midrule
\texttt{NoPos} & $78.2 \pm 0.5\%$ & Permutation Invariant \\
\texttt{ShuffledLabels} & $49.6 \pm 1.2\%$ & Requires $x \to y$ mapping \\
\texttt{FrozenQK} & $49.6 \pm 1.3\%$ & Requires Learned Metric \\
\bottomrule
\end{tabular}
\end{table}

\section{Mechanistic Evidence}

We now investigate how the model implements these statistical decision boundaries. Our analysis reveals that the model does not use a universal algorithm, but adapts its circuit depth to the task geometry.

First, a common hypothesis is that ICL performs nearest-neighbor smoothing~\citep{han2025kernel}. To test this, we compared the model's logits against a Nadaraya-Watson kernel regression estimator. The correlation is weak, confirming that the model is not merely averaging labels based on similarity, but computing a context-dependent sufficient statistic (e.g., centering by $k$). More details are provided in Appendix~\ref{app:kernel_regression}.

\begin{figure}[h]
    \centering
    \includegraphics[width=0.9\linewidth]{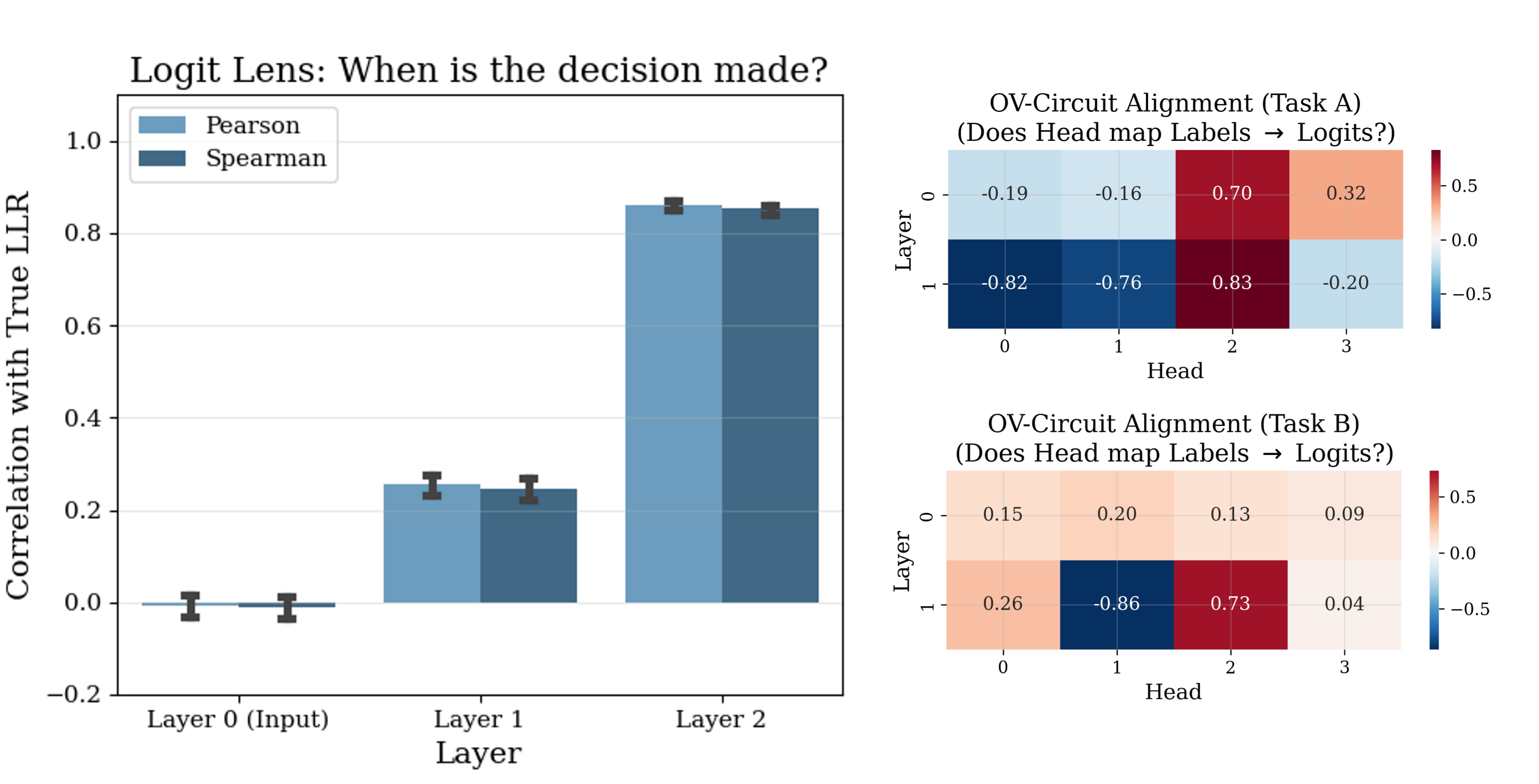}
    \caption{\textbf{Mechanistic Adaptivity.} \textbf{(Left) Logit Lens (Task A):} The correlation with the true LLR rises significantly in Layer 1, suggesting early linear decoding or aggregation. \textbf{(Right) OV Circuit Alignment:} In Task A (top), Layer 0 heads (e.g., Head 2) show strong positive alignment ($>0.7$) with the logit direction, acting as voting ensemble. In Task B (bottom), Layer 0 heads are effectively silent ($<0.26$), implying that the model suppresses early voting to perform deeper sequential processing in Layer 1. Both OV circuits are taken from representative seeds; qualitatively similar behavior persisted across seeds.}
    \label{fig:mech_composite}
\end{figure}

\subsection{Decision Latency and Logit Lens}

Using the Logit Lens technique~\citep{nostalgebraist2020logitlens}, we project intermediate residual states into the vocabulary space. As shown in Figure \ref{fig:mech_composite} (Left), Task A exhibits an early decoding pattern: the representation at Layer 1 shows a partial but decisive correlation with the final target. This suggests that the model is performing a form of preprocessing or summary statistic calculation early in the network which is then refined into a decision. In contrast, nonlinear tasks (Task B) show near-zero correlation until the final layer, indicating a need for deeper composition to estimate energy terms ($\|x\|^2$). See Appendix~\ref{app:logit_lens_task_B}.

\subsection{Hypothesis: Adaptive Circuit Depth and Voting Ensembles}

We find that this decision latency manifests as distinct circuit architectures (Figure \ref{fig:mech_composite}, Right). To interpret the role of individual attention heads, we analyze their Output-Value (OV) circuits~\citep{elhage2021mathematical}. The OV matrix $W_{OV} = W_V W_O$ determines how the features read by a head are projected into the residual stream and, subsequently, the output logits.

In Task A, Layer 0 heads exhibit strong positive alignment ($|\cos \theta| > 0.7$) with the final decision direction. We hypothesize that in this linear regime, the model utilizes a greedy voting ensemble, where heads independently compute partial summary statistics (via forwarding and suppression) that are linearly aggregated to form the decision boundary immediately.

On the other hand, in Task B, Layer 0 heads are effectively silent regarding the decision ($|\cos \theta|$ small). Significant alignment only emerges in Layer 1. This suggests a sequential algorithm where Layer 0 is suppressed or repurposed to compute intermediate features (e.g., squared norms) rather than voting directly.

\section{Conclusion, Limitations, and Future Work}

Importantly, binary hypothesis testing provides a setting where mechanistic interpretability techniques can be compared to a known ground truth. We have demonstrated that toy Transformers trained on dynamic hypothesis testing tasks can approximate the Neyman-Pearson optimal decision rule in-context. By adapting their internal circuit depth (e.g., employing greedy heuristics for linear tasks and sequential processing for nonlinear boundaries) the models recover a sufficient statistic that is highly monotonically correlated with the LLR, matching the performance of a Bayes-optimal oracle in the quadratic regime.

\paragraph{Limitations.} While our controlled synthetic environment allows for exact analytical baselines, it relies on a small two-layer Transformer and relatively low-dimensional Gaussian data. Consequently, it remains an open question to what extent these specific mechanistic behaviors---such as the discrete shift from early voting ensembles to deeper sequential processing---scale to more general statistical tasks or even large language models operating on complex, real-world distributions. Furthermore, our mechanistic interpretability results, including the Logit Lens and OV circuit alignment, establish strong correlational evidence rather than strict causal proofs of the model's internal algorithms. Future work incorporating causal interventions could further substantiate these structural hypotheses.

\paragraph{Future work.} Firstly, conditioning on the in-context dataset reduces each episode to a simple binary hypothesis test, for which the optimal decision rule is characterized by the likelihood-ratio test. A natural extension is to consider composite hypotheses, where class-conditional distributions depend on latent parameters that cannot be eliminated by conditioning alone. In such settings, optimal decision-making requires either marginalization over nuisance parameters or plug-in estimation. Studying ICL in this regime would help distinguish whether models behave more like Bayesian model averaging or approximate maximum-likelihood estimators.

Secondly, our experiments assume balanced class priors and symmetric loss, leading to decision thresholds centered at zero log-likelihood ratio. Extending the framework to asymmetric priors or cost-sensitive objectives would test whether ICL adapts not only the sufficient statistic but also the optimal decision threshold, as prescribed by statistical decision theory.

Finally, binary hypothesis testing provides a minimal setting with sharp optimality guarantees. Extending the analysis to multi-class or sequential testing problems, such as multi-way likelihood-ratio tests or Wald-style sequential procedures, would probe whether ICL can recover more complex decision rules under uncertainty while retaining decision-theoretic interpretability.

\clearpage
\bibliographystyle{iclr2026_conference}
\bibliography{iclr2026_conference}

\clearpage
\appendix
\section{Derivation of Optimal Test Statistics}

For completeness, we derive the analytical LLR for both tasks. Although the marginal problem involves latent task parameters $\phi$, conditioning on the context $C$ renders the hypotheses $H_0$ and $H_1$ simple for each episode. Classical Neyman-Pearson optimality therefore applies at the episode level, and the optimal decision statistic is given by the likelihood ratio conditioned on $C$. The following derivations make this dependence explicit for the two task families considered.

\subsection{Task A: Shifted Mean Discrimination}

Let $\phi = \{\mu, k\}$. The class-conditional distributions are isotropic Gaussians with means $\mu_1 = \mu + k$ and $\mu_0 = -\mu + k$, and covariance $\Sigma = I$. For $x \sim \mathcal{N}(m, I)$,
\begin{equation}
    \log p(x \mid m) = -\frac{d}{2} \log(2\pi) - \frac{1}{2} \|x - m\|^2.
\end{equation}
The LLR is
\begin{align}
    \Lambda(x)
    &= -\frac{1}{2} \|x - (\mu + k)\|^2 + \frac{1}{2} \|x - (-\mu + k)\|^2 \\
    &= 2 \mu^\top x - 2 \mu^\top k.
\end{align}
Thus, the optimal statistic is affine in $\mu^\top (x - k)$; correct classification requires centering with respect to the context-dependent shift.

\subsection{Task B: Variance Discrimination}

For centered Gaussians with variances $\sigma_1^2$ and $\sigma_0^2$,
\begin{equation}
    \log p(x \mid \sigma) = -\frac{d}{2} \log(2\pi \sigma^2) - \frac{\|x\|^2}{2\sigma^2}.
\end{equation}
The LLR is
\begin{align}
    \Lambda(x)
    &= \frac{d}{2} \log \frac{\sigma_0^2}{\sigma_1^2}
    + \frac{\|x\|^2}{2} \left( \frac{1}{\sigma_0^2} - \frac{1}{\sigma_1^2} \right).
\end{align}
The first term is a constant bias, while the data-dependent term is proportional to the energy $\|x\|^2$. Hence, the optimal statistic is purely quadratic.

\section{Experimental Details}

Code, results, and figures are available on \href{https://github.com/farischaudhry/implicit-likelihood-ratio-transformer}{GitHub}.

\subsection{Model Architecture}
We use a toy Transformer architecture designed for set-to-scalar tasks, which we refer to as \texttt{ICLTransformer}.
\begin{itemize}
    \item \textbf{Type:} Bidirectional Transformer Encoder (PyTorch \texttt{nn.TransformerEncoder}).
    \item \textbf{Layers:} 2
    \item \textbf{Attention Heads:} 4
    \item \textbf{Embedding Dimension ($d_{model}$):} 128
    \item \textbf{Feedforward Dimension ($d_{ff}$):} 512
    \item \textbf{Activation:} GELU
    \item \textbf{Normalization:} Post-LayerNorm (\texttt{norm\_first=False})
    \item \textbf{Input Processing:} 
    The input $x \in \mathbb{R}^{16}$ is linearly projected to $d_{model}$. The binary label $y \in \{0, 1\}$ is projected via a separate learnable linear layer. These two projections are combined via element-wise addition to form the final context token embedding, effectively binding the label information to the input features via superposition.
    \item \textbf{Positional Encodings:} Standard learned absolute positional embeddings are added to the sequence.
\end{itemize}

\subsection{Task Specifications}
Data is generated on-the-fly during training. Each batch consists of $B=64$ independent episodes.

\paragraph{Task A: Shifted Mean (Linear).}
\begin{itemize}
    \item \textbf{Input Dimension:} $d_x = 16$.
    \item \textbf{Context Size:} $N = 32$.
    \item \textbf{Latent Parameters:} 
    \begin{itemize}
        \item Discriminative direction $\mu \sim \text{Unif}(\mathbb{S}^{d_x-1})$.
        \item Nuisance shift $k \sim \mathcal{N}(0, \sigma_k^2 I_{d_x})$.
    \end{itemize}
    \item \textbf{Shift Magnitude:} $\sigma_k = 3.0$ (Training), $\sigma_k = 9.0$ (Out-of-distribution (OOD) evaluation).
    \item \textbf{Data Generation:} $x \mid y \sim \mathcal{N}(k + (2y-1)\mu, I)$.
\end{itemize}

\paragraph{Task B: Variance (Nonlinear).}
\begin{itemize}
    \item \textbf{Input Dimension:} $d_x = 16$.
    \item \textbf{Context Size:} $N = 32$.
    \item \textbf{Latent Parameters:} 
    \begin{itemize}
        \item Class 0 Scale $\sigma_0 \sim \text{Unif}[0.5, 3.0]$.
        \item Class 1 Scale $\sigma_1 \sim \text{Unif}[0.5, 3.0]$.
    \end{itemize}
    \item \textbf{Data Generation:} $x \mid y \sim \mathcal{N}(0, \sigma_y^2 I)$.
\end{itemize}

\subsection{Training Hyperparameters}
Models are trained to minimize the Binary Cross Entropy loss on the query label $y_q$.
\begin{itemize}
    \item \textbf{Optimizer:} AdamW ($\beta_1=0.9, \, \beta_2=0.999 $, weight decay $1e-4$).
    \item \textbf{Scheduler:} OneCycleLR.
    \item \textbf{Initial Learning Rate:} $3 \times 10^{-4}$.
    \item \textbf{Batch Size:} 64 tasks per step.
    \item \textbf{Training Duration:} 20 epochs.
    \item \textbf{Seeds:} Accuracy results are reported over 3 random seeds.
\end{itemize}

\subsection{Ablation Variants}

To isolate the mechanism of in-context learning, we evaluated several model variants. Each variant tests a specific hypothesis regarding the inductive bias or information flow required for the task.

\paragraph{Positional Encodings (\texttt{NoPos}, \texttt{FrozenPos}).}
Standard Transformers use positional encodings to process sequences. However, the statistical tasks (shifted mean, variance) are permutation invariant with respect to the context examples. Namely, the learned methodology should be permutation-invariant like sufficent statistics are.
\begin{itemize}
\item \texttt{ICLTransformerNoPos}: 
We completely remove the learned positional embeddings ($P=0$). This tests whether the model treats the context as a set rather than a sequence.
\item \texttt{ICLTransformerFrozenPos}: We initialize positional embeddings randomly but freeze them during training. This tests whether the model requires learned positional information or can utilize random absolute position markers.
\end{itemize}

\paragraph{Attention Mechanism (\texttt{FrozenAttention}, \texttt{FrozenQK}).}
We test whether the attention heads must learn a task-specific metric space or if they can function as random associative memories.
\begin{itemize}
\item \texttt{ICLTransformerFrozenQK}: The Query ($W_Q$) and Key ($W_K$) projections are frozen at initialization. Only the Value ($W_V$) and Output ($W_O$) matrices are trainable. This enforces a fixed, random similarity kernel.
\item \texttt{ICLTransformerFrozenAttention}: All attention weights ($W_Q, W_K, W_V, W_O$) are frozen. Only the feedforward MLPs and embedding projections are trainable.
\end{itemize}

\paragraph{Tokenization Strategy (\texttt{Interleaved}).}
Our default architecture sums input and label embeddings: $e_i = \text{Proj}(x_i) + \text{Proj}(y_i)$, effectively binding the label to the input in a single token.
\begin{itemize}
\item \texttt{ICLTransformerInterleavedEmbeddings}: We replace the bound representation with a standard GPT-style interleaved sequence $[x_1, y_1, x_2, y_2, \dots, x_q]$. This tests whether the additive binding is a necessary inductive bias for efficient learning at this scale ($N=2$ layers).
\end{itemize}

\paragraph{Label Dependence (\texttt{NoLabels}, \texttt{ShuffledLabels}).}
These ablations verify that the model is performing supervised mapping ($x \to y$) rather than unsupervised clustering ($x \to x$).
\begin{itemize}
\item \texttt{ICLTransformerNoLabels}: The context consists only of $x$ vectors; $y$ information is zeroed out.
\item \texttt{ICLTransformerShuffledLabels}: The $y$ labels in the context are randomly permuted within the batch, destroying the specific $x_i \to y_i$ mapping while preserving the marginal distribution of labels.
\item \texttt{ICLTransformerNoisyLabels}: During training, a fraction $p$ of the context labels are flipped ($0 \leftrightarrow 1$). This tests the model's ability to aggregate evidence robustly despite contradictory data points.
\end{itemize}

\section{Supplementary Experimental Results}

\subsection{Task A OOD Generalization Analysis}
\label{app:ood_regression}

To assess whether the model has learned the exact symbolic form of the likelihood ratio or a local approximation, we evaluate it on OOD task where the nuisance shift magnitude $\sigma_k$ is increased from $3.0$ (training) to $9.0$ (validation).

\begin{figure}[b]
    \centering
    \includegraphics[width=0.9\linewidth]{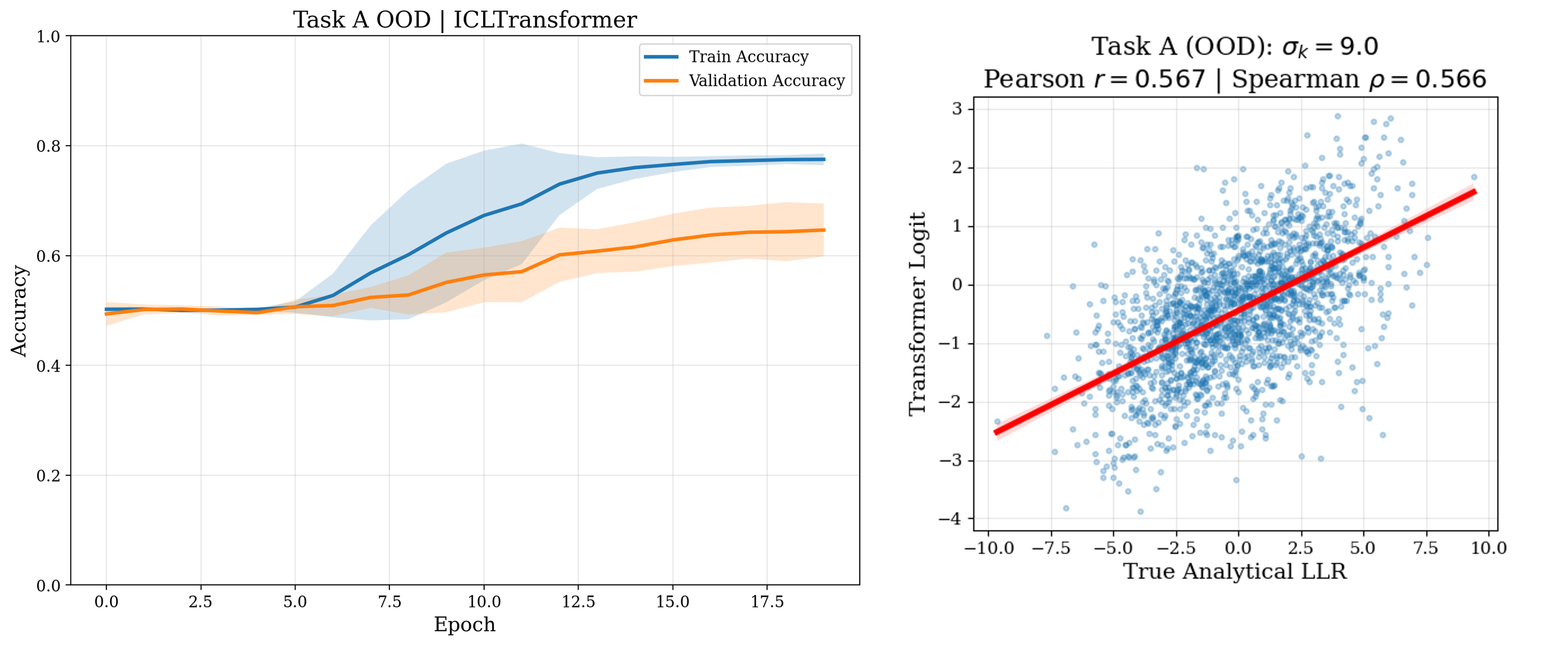}
    \caption{\textbf{OOD Generalization Degradation (Task A).} \textbf{(Left)} Learning curves show a significant generalization gap: while the model masters the training distribution (blue), it struggles to extrapolate to large shifts (orange), achieving only partial generalization. \textbf{(Right)} The correlation with the true LLR degrades to $r=0.567$; the learned decision rule is a local approximation rather than the exact symbolic LLR.}
    \label{fig:ood_composite}
\end{figure}

Figure~\ref{fig:ood_composite} presents the learning dynamics and final decision geometry for this OOD setting.
\begin{itemize}
    \item \textbf{Generalization Gap (Left):} While the training accuracy converges rapidly to $\approx 78\%$ (consistent with the in-distribution baseline), the OOD validation accuracy lags significantly, plateauing at $\approx 64\%$. The delayed rise in validation accuracy suggests a form of partial ``grokking,'' where the model gradually refines its decision rule, but the persistent gap indicates that the learned mechanism does not fully capture the invariant symbolic structure needed for perfect extrapolation.
    \item \textbf{Regression Degradation (Right):} The correlation between the model's logits and the true LLR drops from $r \approx 0.86$ (in-distribution) to $r \approx 0.57$. The increased scatter suggests that the model's internal approximation of the sufficient statistic ($\mu^\top(x-k)$) is calibrated only for the training support and becomes brittle under large shifts.
\end{itemize}

Taken together, these results support the hypothesis that the Transformer implements an amortized approximate inference algorithm: it constructs a decision boundary that mimics the optimal LLR geometry locally, but relies on heuristics that degrade when the task parameters drift far from the training distribution.

\subsection{Full Ablation Results}
\label{app:full_ablation_results}

\begin{table}[H]
\centering
\scriptsize
\caption{\textbf{Full Experimental Results.} We report mean accuracy $\pm$ 95\% CI over 3 seeds for all experimental conditions. The oracle rows represent the theoretical upper bound (Bayes-Optimal Classifier) computed using the true latent task parameters. The model is close to the oracle on Task B, while Task A ablations demonstrate the necessity of learned attention mechanisms.}
\label{tab:full_results_appendix}
\resizebox{\textwidth}{!}{
\begin{tabular}{llcc}
\toprule
\textbf{Experiment / Condition} & \textbf{Model Variant} & \textbf{Train Acc (\%)} & \textbf{Val Acc (\%)} \\
\midrule
\multicolumn{4}{l}{\textit{Theoretical Oracle}} \\
Task A (Shifted Mean) & \texttt{LLR} & --- & $84.6 \pm 1.0$ \\
Task B (Variance) & \texttt{LLR} & --- & $84.0 \pm 1.0$ \\
\midrule
\multicolumn{4}{l}{\textit{Main Tasks}} \\
Task A (Shifted Mean) & \texttt{ICLTransformer} & $77.5 \pm 1.1$ & $78.3 \pm 0.3$ \\
Task B (Variance) & \texttt{ICLTransformer} & $83.0 \pm 0.2$ & $83.0 \pm 0.5$ \\
Task A OOD ($\sigma_k=9.0$) & \texttt{ICLTransformer} & $77.5 \pm 1.1$ & $64.7 \pm 4.8$ \\
\midrule
\multicolumn{4}{l}{\textit{Architecture Ablations (Task A)}} \\
No Positional Encodings & \texttt{NoPos} & $77.5 \pm 1.1$ & $78.2 \pm 0.5$ \\
Frozen Positional Encodings & \texttt{FrozenPos} & $77.5 \pm 1.2$ & $78.1 \pm 0.6$ \\
Frozen Attention Weights & \texttt{FrozenAttention} & $49.9 \pm 0.2$ & $50.4 \pm 0.7$ \\
Frozen Q/K Projections & \texttt{FrozenQK} & $49.7 \pm 0.1$ & $49.6 \pm 1.3$ \\
Interleaved Embeddings ($x, y$) & \texttt{Interleaved} & $49.8 \pm 0.3$ & $49.4 \pm 1.2$ \\
\midrule
\multicolumn{4}{l}{\textit{Data Structure Ablations (Task A)}} \\
Shuffled Context Pairs & \texttt{ShuffledContext} & $77.5 \pm 1.0$ & $78.0 \pm 0.6$ \\
Shuffled Labels Only & \texttt{ShuffledLabels} & $49.8 \pm 0.2$ & $49.6 \pm 1.3$ \\
No Labels & \texttt{NoLabels} & $50.0 \pm 0.1$ & $50.2 \pm 1.6$ \\
Increased Context Size & \texttt{ICLTransformer} & $75.4 \pm 4.7$ & $75.9 \pm 4.3$ \\
\midrule
\multicolumn{4}{l}{\textit{Label Noise Robustness (Task A)}} \\
Noisy Labels ($p=0.1$) & \texttt{NoisyLabels} & $67.7 \pm 11.2$ & $70.2 \pm 11.6$ \\
Noisy Labels ($p=0.2$) & \texttt{NoisyLabels} & $52.1 \pm 2.9$ & $53.3 \pm 5.7$ \\
Noisy Labels ($p=0.4$) & \texttt{NoisyLabels} & $49.7 \pm 0.2$ & $49.7 \pm 1.4$ \\
\bottomrule
\end{tabular}
}
\end{table}

\subsection{Comparison with Kernel Regression}
\label{app:kernel_regression}

\begin{figure}[h]
    \centering
    \includegraphics[width=0.5\linewidth]{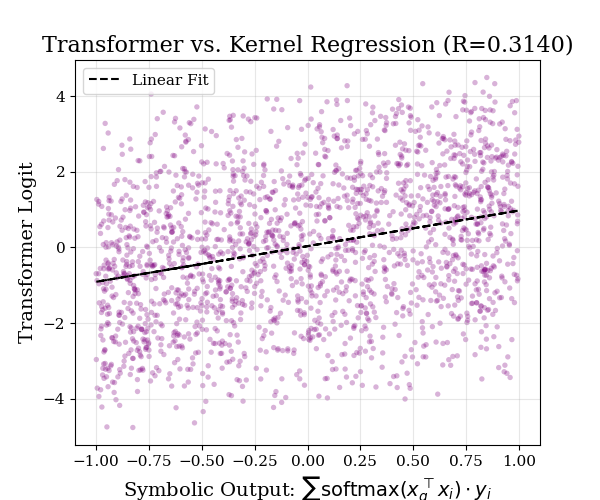}
    \caption{\textbf{Transformer vs. Kernel Regression.} The low correlation indicates the model implements a more complex decision rule than similarity-based label smoothing.}
    \label{fig:kernel_fail}
\end{figure}

To verify that the model is performing algorithmic reasoning rather than simple pattern matching, we compare its outputs to a Nadaraya-Watson~\citep{nadaraya1964regression} estimator using a dot-product kernel:
\begin{equation}
\hat{y}_{KR}(x_q) = \sum_{i=1}^N \frac{e^{x_q^\top x_i}}{\sum_j e^{x_q^\top x_j}} y_i 
\end{equation}
As shown in Figure~\ref{fig:kernel_fail}, the correlation between the Transformer's logits and the Kernel Regression estimator is weak ($\rho \approx 0.33$). This falsifies the hypothesis that the model is merely smoothing labels based on raw input similarity. In Task A, the optimal decision requires computing distances relative to a dynamic shift $k$, which a simple dot product kernel cannot capture without explicit centering.

\subsection{Logit Lens Analysis: Task B}
\label{app:logit_lens_task_B}

In contrast to the linear regime of Task A, where decision-relevant information emerges early in the residual stream, Task B exhibits a delayed decision profile.

\begin{figure}[h]
    \centering
    \includegraphics[width=0.6\linewidth]{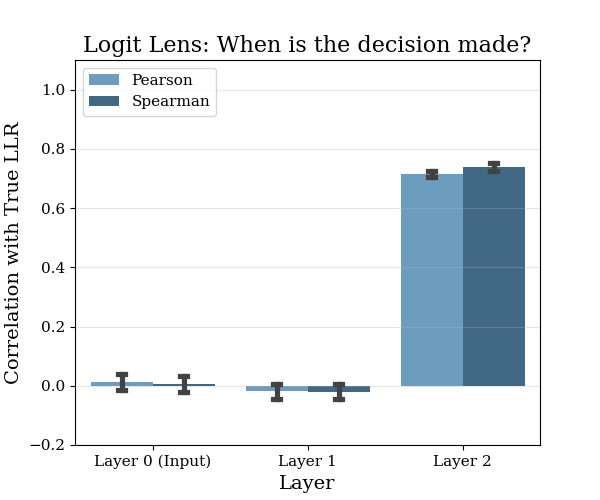}
    \caption{\textbf{Logit Lens for Task B.} The Pearson and Spearman correlations with the true LLR are effectively zero for the initial layers, spiking only at the final output. This confirms that the model does not perform a greedy linear approximation early in the network, but relies on the full depth of the Transformer to construct some nonlinear decision boundary.}
    \label{fig:logit_lens_task_b}
\end{figure}

As illustrated in Figure~\ref{fig:logit_lens_task_b}, the correlation between the intermediate residual states and the LLR remains negligible ($\approx 0$) through Layer 0 and Layer 1. A decisive spike in correlation appears only at the final output stage. This latency supports the hypothesis that nonlinear statistical inference requires a deeper, sequential circuit. We posit that the early layers are occupied with computing the necessary sufficient statistics (e.g., the quadratic energy term $\|x\|^2$) which are geometrically orthogonal to the final linear readout until fully assembled.

\end{document}

%% file: math_commands.tex
\usepackage{amsmath,amsfonts,bm}

\def\eqref#1{equation~\ref{#1}}

\def\1{\bm{1}}

\DeclareMathAlphabet{\mathsfit}{\encodingdefault}{\sfdefault}{m}{sl}
\SetMathAlphabet{\mathsfit}{bold}{\encodingdefault}{\sfdefault}{bx}{n}

